\documentclass{article}
\usepackage{spconf,amssymb,amsmath,epsfig}
\usepackage{microtype}
\usepackage{graphicx}
\usepackage{multirow}
\usepackage{booktabs}
\usepackage{amssymb,amsmath,graphicx}
\usepackage{caption}
\usepackage{subcaption}
\usepackage[export]{adjustbox}
\usepackage{moredefs} % for use of \defcommand. NOTE: can replace \defcommand by \(re)newcommand, but \defcommand will auto reassign an existing command
\usepackage{color}

\defcommand{\vec}[1]{\mathbf{#1}} % vectors in bold instead of with an arrow on top

\def\x{{\mathbf x}}
\def\y{{\mathbf y}}

\def\vx{\mathbf{x}}
\def\vh{\mathbf{h}}

\title{State-of-the-art Speech Recognition \\ With Sequence-to-Sequence Models}
\name{
 Chung-Cheng Chiu,
 Tara N. Sainath,
 Yonghui Wu,
 Rohit Prabhavalkar,
 Patrick Nguyen,
\secondlinename{
 Zhifeng Chen,
 Anjuli Kannan,
 Ron J. Weiss,
 Kanishka Rao,
 Ekaterina Gonina,
}
\thirdlinename{
 Navdeep Jaitly,
 Bo Li,
 Jan Chorowski,
 Michiel Bacchiani
}
\address{Google, USA \\ % Technically LLC.
\fontsize{9}{9}\selectfont\ttfamily\upshape
\{chungchengc,tsainath,yonghui,prabhavalkar,drpng,zhifengc,anjuli\\
\fontsize{9}{9}\selectfont\ttfamily\upshape
  ronw,kanishkarao,kgonina,ndjaitly,boboli,chorowski,michiel\}@google.com}}

\begin{document}
\maketitle
\ninept
\begin{abstract}
 Attention-based encoder-decoder architectures such as Listen, Attend, and
 Spell (LAS), subsume the acoustic, pronunciation and language model
 components of a traditional automatic speech recognition (ASR) system into a
 single neural network.
 In previous work, we have shown that such architectures are
 comparable to state-of-the-art ASR systems on dictation tasks, but it
 was not clear if such architectures would be practical for more
 challenging tasks such as voice search.
 In this work, we explore a variety of structural and optimization
 improvements to our LAS model which significantly improve performance.
 On the structural side, we show that word piece models can be used instead of
 graphemes.
 We also introduce a multi-head attention architecture, which offers
 improvements over the commonly-used single-head attention.
 On the optimization side, we explore synchronous training, scheduled sampling,
 label smoothing, and minimum word error rate optimization, which are all shown
 to improve accuracy.
 We present results with a unidirectional LSTM encoder for streaming
 recognition.
 On a $12,500$~hour voice search task, we find that the proposed changes improve
 the WER from $9.2$\% to $5.6$\%, while the best conventional
 system achieves $6.7$\%; on a dictation task our model achieves a WER of $4.1\%$
 compared to $5\%$ for the conventional system.
\end{abstract}

\section{Introduction\label{sec:intro}}
%Budget: 0.5 page
Sequence-to-sequence models have been gaining in popularity in the automatic speech recognition (ASR) community as a way of folding separate acoustic, pronunciation and language models (AM, PM, LM) of a conventional ASR system into a single neural network. There have been a variety of sequence-to-sequence models explored in the literature, including Recurrent Neural Network Transducer (RNN-T)~\cite{Graves12},  Listen, Attend and Spell (LAS)~\cite{Chan15}, Neural Transducer~\cite{Jaitly16}, Monotonic Alignments~\cite{Colin17} and Recurrent Neural Aligner (RNA)~\cite{Hasim17}. While these models have shown promising results, thus far, it is not clear if such approaches would be practical to unseat the current state-of-the-art, HMM-based neural network acoustic models, which are combined with a separate PM and LM in a conventional system. Such sequence-to-sequence models are fully neural, without finite state transducers, a lexicon, or text normalization modules. Training such models is simpler than conventional ASR systems: they do not require bootstrapping from decision trees or time alignments generated from a separate system. To date, however, none of these models has been able to outperform a state-of-the art ASR system on a large vocabulary continuous speech recognition (LVCSR) task. The goal of this paper is to explore various structure and optimization improvements to allow sequence-to-sequence models to significantly outperform a conventional ASR system on a voice search task.

Since previous work showed that LAS offered improvements over other
sequence-to-sequence models~\cite{RohitSeq17}, we focus on improvements to the
LAS model in this work.  The LAS model is a single neural network that includes
an \emph{encoder} which is analogous to a conventional acoustic model, an
\emph{attender} that acts as an alignment model, and a \emph{decoder} that is
analogous to the language model in a conventional system.  We consider both
modifications to the model structure, as well as in the optimization process. 
On the structure side, first, we explore word piece models (WPM) which have been
applied to machine translation~\cite{Yonghui16} and more recently to speech in
RNN-T~\cite{Rao17} and LAS~\cite{lsd2017iclr}.  We compare graphemes and WPM for
LAS, and find modest improvement with WPM. 
Next, we explore incorporating multi-head attention~\cite{Vaswani17}, which
allows the model to learn to attend to multiple locations of the encoded
features.  Overall, we get 13\% relative improvement in WER with these structure
improvements.
%However, WPM models gives us an additional advantage that we get a much stronger word-level language model on the decoder compared to graphemes. Incorporating an external language model during the beam search \cite{Chorowski17} has shown improvements for a small vocabulary WSJ task. We find that incorporating an external language model during the beam search, gives XX\% relative improvements for WPM compared to graphemes, which gives YY\%.

On the optimization side, we explore a variety of strategies as well. 
Conventional ASR systems benefit from discriminative sequence training, which
optimizes criteria more closely related to WER~\cite{bedk:nn}.  Therefore, in
the present work, we explore training our LAS models to minimize the number of
expected word errors (MWER)~\cite{RohitMBR18}, which significantly improves
performance.  Second, we include scheduled sampling
(SS)~\cite{BengioVinyalsJaitlyEtAl15,Chan15}, which feeds the previous label
prediction during training rather than ground truth.  Third, label
smoothing~\cite{Szegedy16} helps to make the model less confident in its
predictions, and is a regularization mechanism that has successfully been
applied in both vision~\cite{Szegedy16} and speech tasks~\cite{Jan15,Jan17}. 
Fourth, while many of our models are trained with asynchronous
SGD~\cite{Dean12}, synchronous training has recently been shown to improve
neural systems~\cite{Goyal17}.  We find that all four optimization strategies
allow for additional $27.5$\% relative improvement in WER on top of our
structure improvements.

Finally, we incorporate a language model to rescore N-best lists in the second
pass, which results in a further $3.4$\% relative improvement in WER.  Taken
together, the improvements in model structure and optimization, along with
second-pass rescoring, allow us to improve a single-head attention, grapheme LAS
system, from a WER of $9.2$\% to a WER of $5.6$\% on a voice search task. 
This provides a $16$\% relative reduction in WER compared to a strong
conventional model baseline which achieves a WER of $6.7$\%.  We also observe a
similar trend on a dictation task.
%We also test both models, trained only with the voice search data, on a dictation task that consist of longer utterances.  Our model achieve $4.1\%$ and the conventional model achieve $5\%$ WER, which provides a $18\%$ relative reduction.

In Sections~\ref{ssec:wpm} and~\ref{ssec:lm}, we show how language models can be 
 integrated. Section~\ref{ssec:multihead} further extends the model to multi-head attention.
We explore discriminative training in Sections~\ref{ssec:mbr} and~\ref{ssec:schedsamp}, and
 synchronous training regimes in Section~\ref{ssec:sync}. We use unidirectional encoders for low-latency streaming decoding.
%Unsolved problems: unidirectional, LM integration, low-latency.

%\input{related}
\section{System Overview}
In this section, we detail various structure and optimization improvements to the basic LAS model.
\subsection{Basic LAS Model}
\begin{figure}[h!]
\centering
  \includegraphics[scale=0.30]{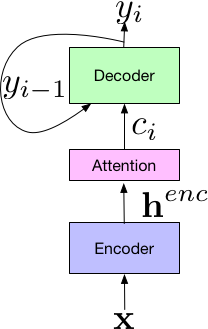}
  \caption{Components of the LAS end-to-end model.}
  \label{fig:las}
\end{figure}
The basic LAS model, used for experiments in this work, consists of 3 modules as shown in Figure~\ref{fig:las}. The \emph{listener} encoder module, which is similar to a standard acoustic model, takes the input features, $\vx$, and maps them to a higher-level feature representation, $\vh^{enc}$.
The output of the encoder is passed to an \emph{attender}, which determines which encoder features in $\vh^{enc}$ should be attended to in order to predict the next output symbol, $y_i$, similar to a dynamic time warping (DTW) alignment module. Finally, the output of the attention module is passed to the \emph{speller} (i.e., decoder), which takes the attention context, $c_i$, generated from the attender, as well as an embedding of the previous prediction, $y_{i-1}$, in order to produce a probability distribution, $P(y_i|y_{i-1}, \ldots, y_0, \vx)$, over the current sub-word unit, $y_i$, given the previous units, $\{y_{i-1}, \ldots, y_0\}$, and input, $\vx$.
%We can think of the decoder as similar to a language model.

\subsection{Structure Improvements}
\subsubsection{Wordpiece models\label{ssec:wpm}}
Traditionally, sequence-to-sequence models have used graphemes (characters) as output units, as this folds the AM, PM and LM into one neural network, and side-steps the problem of out-of-vocabulary words~\cite{Chan15}.
Alternatively, one could use longer units such as word pieces or shorter units such as context-independent phonemes~\cite{TaraCip2018}. One of the disadvantages of using phonemes is that it requires having an additional PM and LM, and was not found to improve over graphemes in our experiments~\cite{TaraCip2018}.

Our motivation for looking at word piece models (WPM) is a follows. Typically, word-level LMs have a much lower perplexity compared to grapheme-level LMs~\cite{Anjuli18}. Thus, we feel that modeling word pieces allows for a much stronger decoder LM compared to graphemes.  In addition, modeling longer units improves the effective memory of the decoder LSTMs, and allows the model to potentially memorize pronunciations for frequently occurring words. Furthermore, longer units require fewer decoding steps; this speeds up inference in these models significantly. Finally, WPMs have shown good performance for other sequence-to-sequence models such as RNN-T~\cite{Rao17}.

The word piece models~\cite{wordpiece_schuster} used in this paper are sub-word units, ranging from graphemes all the way up to entire words. Thus, there are
no out-of-vocabulary words with word piece models. The word piece models are trained
to maximize the language model likelihood over the training set. As in~\cite{Yonghui16}, the word pieces are
``position-dependent'', in that a special word separator marker is used to denote
word boundaries. Words are segmented deterministically and independent of context, using a greedy algorithm.
%% The training transcripts are segmented with the word piece algorithm (greedy
%% longest prefix sequence). Thus, the word piece decomposition is independent of the audio, and
%% the context in which a word occurs.

%While word piece models correspond to variable durations, it is expected that the multi-head
%attention model would be necessary to modulate the attention duration. We did not find this
%to be the case.

\subsubsection{Multi-headed attention\label{ssec:multihead}}
\begin{figure}[t]
\centering
  \includegraphics[scale=0.35]{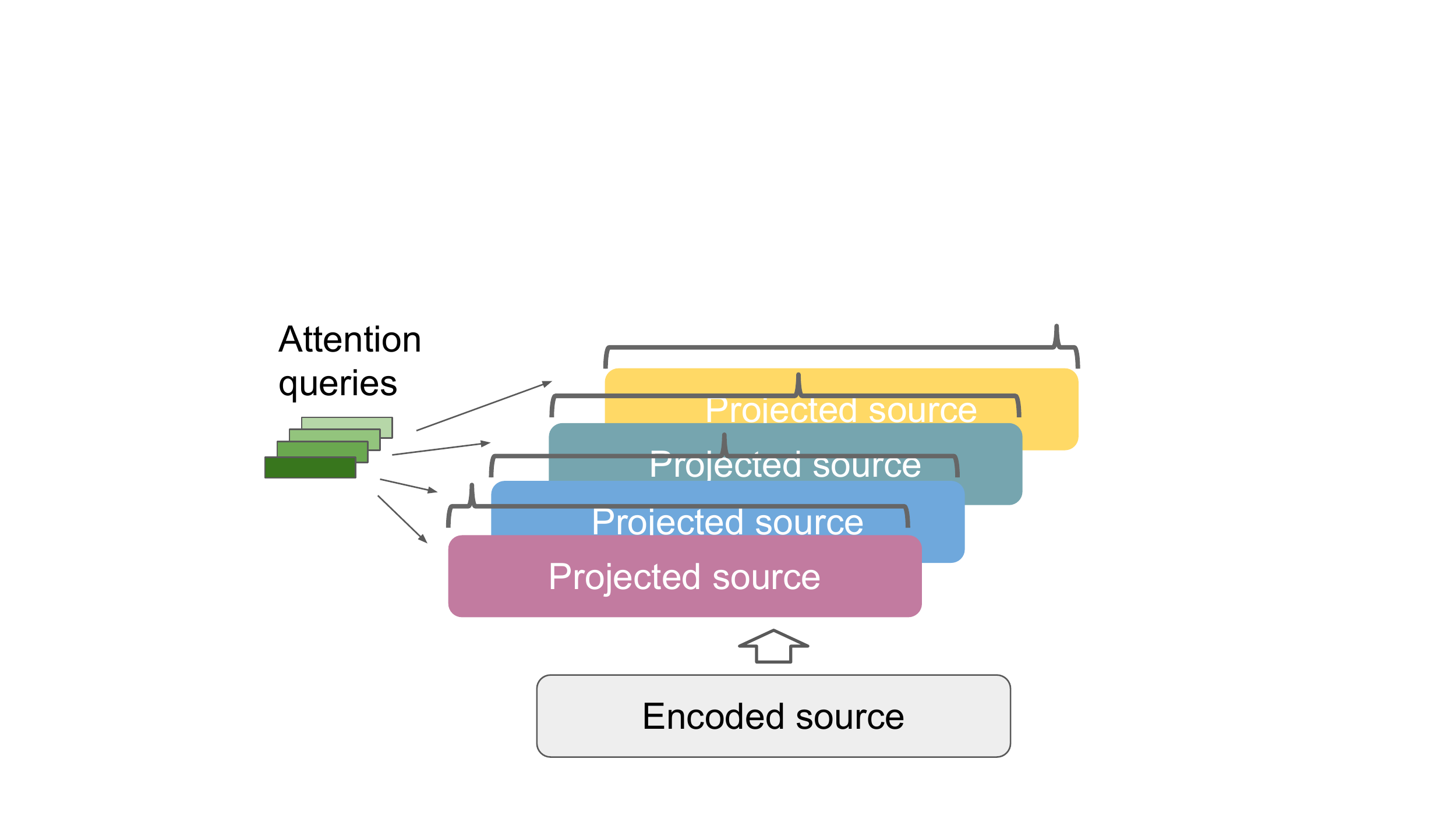}
  \caption{Multi-headed attention mechanism.}
  \label{fig:mha}
\end{figure}
Multi-head attention (MHA) was first explored in~\cite{Vaswani17} for machine translation, and we extend this work to explore the value of MHA for speech.  Specifically, as shown in Figure~\ref{fig:mha}, MHA extends the conventional attention mechanism to have multiple heads, where each head can generate a different attention distribution.  This allows each head to have a different role on attending the encoder output, which we hypothesize makes it easier for the decoder to learn to retrieve information from the encoder.  In the conventional, single-headed architecture, the model relies more on the encoder to provide clearer signals about the utterances so that the decoder can pickup the information with attention.   We observed that MHA tends to allocate one of the attention head to the beginning of the utterance which contains mostly background noise, and thus we hypothesize MHA better distinguish speech from noise when the encoded representation is less ideal, for example, in degraded acoustic conditions.
%\begin{itemize}
%\item Multi-head vs. single head
%\item Figure to illustrate multi-headed attention
%\item Example attention plot
%\item Can be efficiently implemented by projecting encoder output
%\end{itemize}

\subsection{Optimization improvements\label{sec:opt-improvements}}
%In this section, we describe various optimization improvements explored.
\subsubsection{Minimum Word Error Rate (MWER) Training\label{ssec:mbr}}
Conventional ASR systems are often trained to optimize a sequence-level
criterion (e.g., state-level minimum Bayes risk (sMBR)~\cite{bedk:nn}) in
addition to CE or CTC training. Although the loss function that we optimize for the attention-based systems is
\emph{a sequence-level loss function}, it is not closely related to
the metric that we actually care about, namely word error rate. There have been a variety of methods explored in the literature to address this issue in the context of sequence-to-sequence models~\cite{Bahdanau16,RanzatoChopraAuliEtAl16,Hasim17,RohitMBR18}.
In this work, we focus on the minimum expected word error rate (MWER) training
that we proposed in~\cite{RohitMBR18}.

In the MWER strategy, our objective is to minimize the expected number of word errors.
The loss functon is given by Equation~\ref{eq:seq1}, where
$\mathcal{W}(\mathbf{y, \mathbf{y}^*})$ denotes the number of word errors in the
hypothesis, $\mathbf{y}$, compared to the ground-truth label sequence,
$\mathbf{y}^*$.
This first term is interpolated with the standard cross-entropy based loss,
which we find is important in order to stabilize
training~\cite{RohitMBR18,povey2002minimum}.
\begin{equation}
	\mathcal{L}_\text{MWER} = \mathbb{E}_{P(\mathbf{y}|\mathbf{x})}
	        [\mathcal{W}(\mathbf{y, \mathbf{y}^{*}})] + \lambda \mathcal{L}_\text{CE}
               \label{eq:seq1}
\end{equation}
The expectation in Equation~\ref{eq:seq1} can be approximated either via
sampling~\cite{Hasim17} or by restricting the summation to an N-best list of
decoded hypotheses as is commonly used for sequence training~\cite{bedk:nn};
the latter is found to be more effective in our experiments~\cite{RohitMBR18}.
We denote by $\text{NBest}(\x, N) = \{\y_1, \cdots, \y_N\}$, the
set of N-best hypotheses computed using beam-search
decoding~\cite{SutskeverVinyalsLe14} for the input utterance $\x$.
The loss function can then be approximated as shown in
Equation~\ref{eq:seq2}, which weights the \emph{normalized} word errors from
each hypothesis by the probability $\widehat{P}(\mathbf{y}_i|\mathbf{x})$
concentrated on it:
\begin{equation}
	\mathcal{L}^\text{N-best}_\text{MWER} = \frac{1}{N} \sum_{y_i \in
	\text{NBest}(\x, N)}
	[\mathcal{W}(\mathbf{y_i, \mathbf{y}^{*}}) - \widehat{\mathcal{W}}] \widehat{P}(\mathbf{y}_i | \mathbf{x}) + \lambda \mathcal{L}_\text{CE}
        \label{eq:seq2}
\end{equation}
\noindent
where, $\widehat{P}(\y_i|\x) = \frac{P(\y_i|\x)}{\sum_{\y_i \in \text{NBest}(\x,
N)} P(\y_i|\x)}$, represents the distribution re-normalized over just the N-best
hypotheses, and $\widehat{\mathcal{W}}$ is the average number of word errors
over all hypotheses in the N-best list.
Subtracting $\widehat{\mathcal{W}}$ is applied as a form of variance reduction
since it does not affect the gradient~\cite{RohitMBR18}.
%MWER training was found to give between a 7-8\% relative improvement in WER over a cross-entropy trained LAS system in \cite{RohitMBR18}.

\subsubsection{Scheduled Sampling\label{ssec:schedsamp}}
We explore scheduled sampling~\cite{BengioVinyalsJaitlyEtAl15} for training the decoder.  Feeding the ground-truth label as the previous prediction (so-called {\em teacher forcing}) helps the decoder to learn quickly at the beginning, but introduces a mismatch between training and inference.  The scheduled sampling process, on the other hand, samples from the probability distribution of the previous prediction (i.e., from the softmax output) and then uses the resulting token to feed as the previous token when predicting the next label.  This process helps reduce the gap between training and inference behavior.  Our training process uses teacher forcing at the beginning of training steps, and as training proceeds, we linearly ramp up the probability of sampling from the model's prediction to $0.4$ at the specified step, which we then keep constant until the end of training. The step at which we ramp up the probability to $0.4$ is set to $1$~million steps and $100,000$~steps for asynchronous and synchronous training respectively (See section~\ref{ssec:sync}).
%Sampling rate ramp up and its effect on per-step accuracy

\subsubsection{Asynchronous and Synchronous Training\label{ssec:sync}}
We compare both asynchronous~\cite{Dean12} and synchronous training~\cite{Goyal17}. As shown in~\cite{Goyal17}, synchronous training can potentially provide faster convergence rates and better model quality, but also requires more effort in order to stabilize network training. Both approaches have a high gradient variance at the beginning of the training when using multiple replicas~\cite{Dean12}, and we explore different techniques to reduce this variance.  In asynchronous training we use replica ramp up: that is, the system will not start all training replicas at once, but instead start them gradually. In synchronous training we use two techniques: learning rate ramp up and a gradient norm tracker.  The learning rate ramp up starts with the learning rate at $0$ and gradually increases the learning rate, providing a similar effect to replica ramp up.  The gradient norm tracker keeps track of the moving average of the gradient norm, and discards gradients with significantly higher variance than the moving average.  Both approaches are crucial for making synchronous training stable.

\subsubsection{Label smoothing\label{ssec:labelsmooth}}
Label smoothing~\cite{Szegedy16,Jan17} is a regularization mechanism to prevent the model from making over-confident predictions.  It encourages the model to have higher entropy at its prediction, and therefore makes the model more adaptable.  We followed the same design as~\cite{Szegedy16} by smoothing the ground-truth label distribution with a uniform distribution over all labels.

\subsection{Second-Pass Rescoring\label{ssec:lm}}
%% While the LAS decoder is itself already a
%% neural language model (LM), the model is only exposed to training transcripts.
While the LAS decoder topology is that of
neural language model (LM), it can function as a language model; but it is only exposed to training transcripts.
An external LM, on the other hand, can leverage large amounts of
additional data for which we only have text (no audio). To address the
potentially weak LM learned by the decoder,
we incorporate an external LM during inference only.

The external LM is a large 5-gram LM trained on text data from a variety
of domains.  Since domains have different predictive value for our LVCSR
task, domain-specific LMs are first trained, then combined together
using Bayesian-interpolation~\cite{Allauzen2011}.% \textbf{AK: Are there any
%other citations for prod LM besides Allauzen+Riley 2011} \TS{anjuli - ask montse or EJ about this}

We incorporate the LM in the second-pass by means of log-linear interpolation.
In particular, given the N-best
hypotheses produced by the LAS model via beam search, we determine the final
transcript $\mathbf{y}^*$ as:
\begin{equation}
\mathbf{y}^* = \underset{\y}{\arg\max} \log P(\y|\x) + \lambda \log P_{LM}(\y) + \gamma \text{len}(\y)
\end{equation}
\noindent
where, $P_{LM}$ is provided by the LM, $\text{len}(\y)$ is the number of words in $\y$,
and $\lambda$ and $\gamma$ are tuned on a development set.
Using this criterion, transcripts which have a low language model
probability will be demoted in the final ranked list.  Additionally, the
the last term  addresses the common observation that
the incorporation of an LM leads to a higher rate of deletions.

\section{Experimental Details \label{sec:experiments}}
Our experiments are conducted on a $\sim$12,500 hour training set consisting of
15 million English utterances.
The training utterances are anonymized and hand-transcribed, and are
representative of Google's voice search traffic.
This data set is created by artificially corrupting clean utterances using a
room simulator, adding varying degrees of noise and reverberation such that the
overall SNR is between 0dB and 30dB, with an average SNR of 12dB.
The noise sources are from YouTube and daily life noisy environmental
recordings.  We report results on a set of $\sim$14.8K utterances extracted from Google
traffic, and also evaluate the resulting model, trained with only voice search
data, on a set of 15.7K dictation utterances that have longer sentences
than the voice search utterances.

All experiments use 80-dimensional log-Mel features, computed with a 25ms
window and shifted every 10ms.  Similar to~\cite{Hasim15, Golan16}, at the current frame, $t$, these features are stacked with 3 frames to the left and downsampled to a 30ms frame rate.  The encoder network architecture consists of 5 long short-term memory~\cite{HochreiterSchmidhuber97} (LSTM) layers.  We explored both unidirectional~\cite{Schuster97} and bidirectional LSTMs, where the unidirectional LSTMs have $1,400$ hidden units and bidirectional LSTMs have $1,024$ hidden units in each direction ($2,048$ per layer).  Unless otherwise stated, experiments are reported with unidirectional encoders. Additive attention~\cite{Bahdanau15} is used for both single-headed and multi-headed attention experiments.  All multi-headed attention experiments use 4 heads.  The decoder network is a 2-layer LSTM with $1,024$ hidden units per layer.

All neural networks are trained with the cross-entropy criterion (which is used
to initialize MWER training) and are trained using
TensorFlow~\cite{AbadiAgarwalBarhamEtAl15}.

\section{Results\label{sec:results}}
\subsection{Structure Improvements}
Our first set of experiments explore different structure improvements to the LAS model. Table~\ref{tab:results_alg} compares performance for LAS models given graphemes (\texttt{E1}) and WPM (\texttt{E2}). The table indicates that WPM perform slightly better than graphemes. This is consistent with the finding in~\cite{Rao17} that WPM provides a stronger decoder LM compared to graphemes, resulting in roughly a $2$\% relative improvement in WER (WERR).
%However, for bidi models the performance of both units is roughly the same. It has been shown in \TS{cite anjuli paper} that the gains by incorporating an external LM is less with a bidirectional model compared to a unidirectional model. Since a WPM model provides a stronger LM compared to graphemes, we hypothesize this is why WPM and bidirecitonal models does not provide gains over graphemes.

Second, we compare the performance of MHA with WPM, as shown by experiment \texttt{E3} in the table. We see MHA provides around a $11.1$\% improvement. This indicates that having the model focus on multiple points of attention in the input signal, which is similar in spirit to having a language model passed from the encoder, helps significantly. Since models with MHA and WPM perform best, we explore the proposed optimization methods on top of this model in the rest of the paper.
\begin{table}[h!]
  \centering
  \begin{tabular}{|l|l|c|c|}\hline
    Exp-ID    & Model    & WER & WERR \\\hline
 \texttt{E1}  & Grapheme & 9.2   & - \\
 \texttt{E2}  & WPM      & 9.0   & 2.2\%\\
 \texttt{E3}  & + MHA    & 8.0   & 11.1\%\\
 \hline
  \end{tabular}
  \caption{Impact of word piece models and multi-head attention.}
  \label{tab:results_alg}
\end{table}
\vspace{-0.3in}

\subsection{Optimization improvements}
We explore the performance of various optimization improvements discussed in Section~\ref{sec:opt-improvements}. Table~\ref{tab:results_opt} shows that including synchronous training (\texttt{E4}) on top of the WPM+MHA model provides a $3.8$\% improvement. Furthermore, including scheduled sampling (\texttt{E5}) gives an additional $7.8$\% relative improvement in WER; label smoothing gives an additional $5.6$\% relative improvement. Finally, MWER training provides $13.4$\%. Overall, the gain from optimizations is around $27.5$\%, moving the WER from $8.0$\% to $5.8$\%.

We see that synchronous training, in our configuration, yields a better converged optimum at similar wall clock time. Interestingly, while scheduled
sampling and minimum word error rate are both discriminative methods, we have observed that their combination continues to yield additive improvements.
Finally, regularization with label smoothing, even with large amounts of data, is proven to be beneficial.
\begin{table}[h!]
  \centering
  \begin{tabular}{|l|l|c|c|}\hline
    Exp-ID  & Model &      WER    & WERR \\\hline
    \texttt{E2} & WPM     & 9.0     & - \\
    \texttt{E3} & + MHA   & 8.0     & 11.1\% \\ \hline
    \texttt{E4} & + Sync  & 7.7     & 3.8\% \\
    \texttt{E5} & + SS    & 7.1     & 7.8\%\\
    \texttt{E6} & + LS    & 6.7     & 5.6\% \\
    \texttt{E7} & + MWER  & 5.8     & 13.4\%\\
 \hline
  \end{tabular}
  \caption{Sync training, scheduled sampling (SS), label smoothing (LS) and minimum word error rate (MWER) training improvements.}
  \label{tab:results_opt}
\end{table}
\vspace{-0.3in}

\subsection{Incorporating Second-Pass Rescoring}
Next, we incorporate second-pass rescoring into our model.
As can be see in Table~\ref{tab:results_lm}, second-pass rescoring improves the WER by $3.4$\%,
from $5.8$\% to $5.6$\%.
\begin{table}[h!]
  \centering
  \begin{tabular}{|l|l|c|}\hline
    Exp-ID  & Model &    WER \\\hline
    \texttt{E7} & WPM + MHA + Sync + SS + LS + MWER & 5.8\\
    \texttt{E8} & + LM & \textbf{5.6} \\ \hline
  \end{tabular}
  \caption{In second pass rescoring, the log-linear combination with a larger LM results in a $0.2$\% WER improvement.}
  \label{tab:results_lm}
\end{table}
\vspace{-0.1in}

\subsection{Unidirectional vs. Bidirectional Encoders}
Now that we have established the improvements from structure, optimization and LM strategies, in this section we compare the gains on a unidirectional and bidirectional systems. Table~\ref{tab:results_bidi} shows that the proposed changes give a $37.8$\% relative reduction in WER for a unidirectional system, while a slightly smaller improvement of $28.4$\% for a bidirectional system. This illustrates that most proposed methods offer improvements independent of model topology.
\begin{table}[h!]
  \centering
  \begin{tabular}{|l|l|c|c|}\hline
    Exp-ID  & Model &    Unidi & Bidi \\\hline
    \texttt{E2} & WPM & 9.0  &  7.4 \\
    \texttt{E8} & WPM + all& \textbf{5.6} & \textbf{5.3} \\ \hline
    WERR & - & 37.8\% & 28.4\% \\ \hline
  \end{tabular}
  \caption{Both unidirectional and bidirectional models benefit from cumulative improvements.}
  \label{tab:results_bidi}
\end{table}

\vspace{-0.2in}
\subsection{Comparison with the Conventional System}
Finally, we compare the proposed LAS model in \texttt{E8} to a state-of-the-art, discriminatively sequence-trained low frame rate (LFR) system~\cite{Golan16} in terms of WER. Table~\ref{tab:wer_prod} shows the proposed sequence-to-sequence model (\texttt{E8}) offers a $16\%$ and $18\%$ relative improvement in WER over our production system (\texttt{E9}) on voice search (VS) and dictation (D) task respectively.  Furthermore, comparing the size of the first-pass models, the LAS model is around 18 times smaller than the conventional model. It is important to note that the second pass model is 80 GB and still dominates model size.
\begin{table}[h!]
  \centering
  \begin{tabular}{|l|l|c|c|}\hline
    Exp-ID      & Model        & VS/D & 1st pass Model Size \\\hline
    \texttt{E8} & Proposed     & \textbf{5.6/4.1} & \textbf{0.4 GB} \\
    \texttt{E9} & Conventional & 6.7/5.0          & 0.1 GB (AM) + 2.2 GB (PM) \\
                & $\,\,$ LFR system &             & + 4.9 GB (LM) = 7.2GB\\
 \hline
  \end{tabular}
  \caption{%Comparison of LAS and conventional LFR model. Both models use second-pass rescoring.
    Resulting WER on voice search (VS)/dictation (D).  The improved LAS outperforms
    the conventional LFR system while being more compact.  Both models use
    second-pass rescoring.
  }
  \label{tab:wer_prod}
\end{table}
\vspace{-0.2in}

\section{Conclusion\label{sec:conclusion}}
We designed an attention-based model for sequence-to-sequence speech recognition. The model integrates acoustic, pronunciation, and language models into a single neural network, and does not require a lexicon or a separate text normalization component. We explored various structure and optimization mechanisms for improving the model.
Cumulatively, structure improvements (WPM, MHA) yielded an $11\%$ improvement in WER, while optimization improvements (MWER, SS, LS and synchronous training) yielded a further $27.5\%$ improvement, and the language model rescoring yielded another $3.4\%$ improvement.  Applied on a Google Voice Search task, we achieve a WER of $5.6\%$, while a hybrid HMM-LSTM system achieves $6.7\%$ WER.  Tested the same models on a dictation task, our model achieves $4.1\%$ and the hybrid system achieves $5\%$ WER.  We note however, that the unidirectional LAS system has the limitation that the entire utterance must be seen by the encoder, before any labels can be decoded (although, we encode the utterance in a streaming fashion). Therefore, an important next step is to revise this model with an streaming attention-based model, such as Neural Transducer~\cite{NeuralTransducer18}.

\bibliographystyle{IEEEbib}
\bibliography{main}
\end{document}